\pdfoutput=1

\documentclass[11pt]{article}

\usepackage[]{ACL2023}

\usepackage{times}
\usepackage{latexsym}

\usepackage[T1]{fontenc}

\usepackage[utf8]{inputenc}

\usepackage{microtype}

\usepackage{inconsolata}

\usepackage{graphicx}
\usepackage{amssymb}

\usepackage{makecell}
\usepackage{amsmath}
\usepackage{subfigure}
\usepackage{hyperref}
\usepackage{booktabs} 
\usepackage{multirow} 
\usepackage{float} 
\usepackage{color}
\usepackage{xcolor} 

\usepackage{colortbl}
\definecolor{GREEN}{RGB}{84,130,53}
\newcommand{\coloritt}{\cellcolor{gray!15}}

\usepackage{amssymb} 
\usepackage{pifont}
\newcommand{\cmark}{\ding{51}}  
\newcommand{\xmark}{\ding{55}}

\title{Assessing Knowledge Editing in Language Models via Relation Perspective}

\author{
   Yifan Wei$^{1,2}$\thanks{~~Equal Contributions.}~  \quad Xiaoyan Yu$^{2,3*}$ \quad \textbf{Huanhuan Ma$^{1,2}$} \quad  \textbf{Fangyu Lei$^{1,2}$} \\
     \textbf{Yixuan Weng$^{2}$}  
     \quad \textbf{Ran Song$^{4}$}  \quad 
  \textbf{Kang Liu$^{1,2}$} 
  \\
  $^{1}$University of Chinese Academy of Sciences 
  $^{2}$Institute of Automation, CAS\\
  $^{3}$School of Computer Science and Technology, Beijing Institute of Technology \\
  $^{4}$Kunming University of Science and Technology\\
  \texttt{xiaoyan.yu@bit.edu.cn},
  \texttt{\{weiyifan2021,mahuanhuan2021,leifangyu2022\}@ia.ac.cn} \\
  \texttt{wengsyx@gmail.com, song\_ransr@163.com, kliu@nlpr.ia.ac.cn}
}

\begin{document}
\maketitle
\begin{abstract}

Knowledge Editing (KE) for modifying factual knowledge in Large Language Models (LLMs) has been receiving increasing attention. 
However, existing knowledge editing methods are entity-centric, and it is unclear whether this approach is suitable for a relation-centric perspective.
To address this gap, this paper constructs a new benchmark named \textbf{RaKE}, which 
focuses on \textbf{R}elation b\textbf{a}sed \textbf{K}nowledge \textbf{E}diting.
In this paper, we establish a suite of innovative metrics for evaluation and conduct comprehensive experiments involving various knowledge editing baselines. 
We notice that existing knowledge editing methods exhibit the potential difficulty in their ability to edit relations. Therefore, we further explore the role of relations in factual triplets within the transformer.
Our research results confirm that knowledge related to relations is not only stored in the FFN network but also in the attention layers. This provides experimental support for future relation-based knowledge editing methods.

\end{abstract}

\section{Introduction}
    
    Large Language Models (LLMs), trained on large-scale knowledge corpora such as Wikipedia, exhibit remarkable performance across various natural language processing tasks \citep{ma2023ex, lei2023tableqakit}.
    However, current LLMs face challenges posed by errors, biases, and inappropriate information \cite{neeman2022disentqa,guo2022auto}.
    Meanwhile, LLMs need to adapt to emerging knowledge over time and eliminate outdated knowledge \cite{kasai2022realtime,wei2023menatqa}.
    To maintain the accuracy and reliability of LLMs, the task of Knowledge Editing (KE)\footnote{In this paper, the term "knowledge editing" is equivalent to "model editing" and "memory editing".}, which involves modifying and updating the internal knowledge of language models, has recently gained significant attention.
    
    \begin{figure}[t]
       \begin{center}
       \includegraphics[width=1\linewidth]{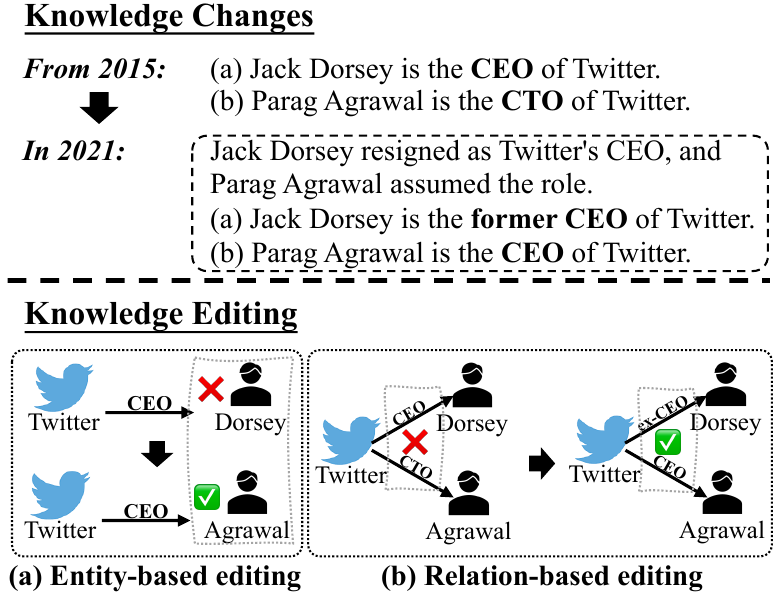}
       \end{center}
       \caption{
       As time progresses, relationships between entities undergo continuous changes. 
       In real-world scenarios, such as Wikipedia, updating factual knowledge sometimes necessitates the modification of relationships to accurately reflect evolving information.
       }
       \label{fig:example}
    \end{figure}
    
    The factual knowledge encapsulated in language models can be represented as the relation between subject and object in the form of $(s, r, o)$\footnote{Knowledge triples: (subject entity, relation, object entity).}. 
    As time progresses, the relations between entities also undergo changes, 
    as illustrated in Figure \ref{fig:example} (b).
    For instance, consider the evolution of Parag Agrawal's role at Twitter\footnote{\url{https://en.wikipedia.org/wiki/Twitter,_Inc.}}: 
    \textit{“From 2015, Parag Agrawal is the \textbf{CTO} of Twitter,”} transforms into 
    \textit{“In 2021, Parag Agrawal is the \textbf{CEO} of Twitter.”}
    The intuitive need arises to directly modify the relation (“CTO” to “CEO”) to accurately reflect this evolving knowledge. 
    However, existing attempts focuses on editing from the entity perspective \cite{de2021editing,mitchell2021fast,mitchell2022memory,dong2022calibrating,huang2023transformer,dai2021knowledge,meng2022locating,meng2022mass,zheng2023can,zhong2023mquake}, ignoring the modification of factual knowledge from the relation perspective.

    To fill this gap, we construct a \textbf{R}elation-b\textbf{a}sed \textbf{K}nowledge \textbf{E}diting benchmark called \textbf{RaKE}, and extend previous evaluation principles \citep{mitchell2021fast,elazar2021measuring} 
    to the perspective of relation.
    Then, we empirically investigate the outcomes of existing methods on relation-based editing.
    Surprisingly, the experimental results reveal that relation-based editing lags far behind entity-based editing, even though they should ideally be consistent since the original and the altered triples are the same.
    To delve into the reasons causing such inconsistency, we conducted a causal tracing analysis on the relation $r$ within the knowledge triple $(s, r, o)$ and investigated how and where the relation memories are stored in LLMs.
    The results show evidence that the relation memories are not only related to the feed-forward network (FFN) but also to the attention layer.
    Due to the fact that entity-based methods primarily modify parameters within the feed-forward network (FFN), our experiments indicate that the underperformance of current relation-based editing stems from a lack of modification to knowledge neurons associated with the attention layer. 
    We hope that our work can provide the NLP community with insights.\footnote{The dataset and code are released in \href{https://github.com/weiyifan1023/Knowledge-Edit-based-on-Relation-Perspective}{https://github.com/weiyifan/Knowledge-Editing} }
    
    Our main contributions are summarized as follows:
    \begin{itemize}
        \item For the first time, we identify the importance of knowledge editing from a relational perspective and construct a new benchmark, RaKE, tailored for relation-based editing.
        \item We conducted extensive experiments using various baseline methods, and the results reveal significant limitations in the current approaches to relation-based editing.
        \item Our results confirm the crucial role of not only the feed-forward network but also the attention modules in storing relational knowledge.
        This insight provides valuable guidance for future KE methods.
    \end{itemize}

\section{Preliminaries}
    
\begin{figure*}[t]
        \begin{center}
        \includegraphics[width=1\linewidth]{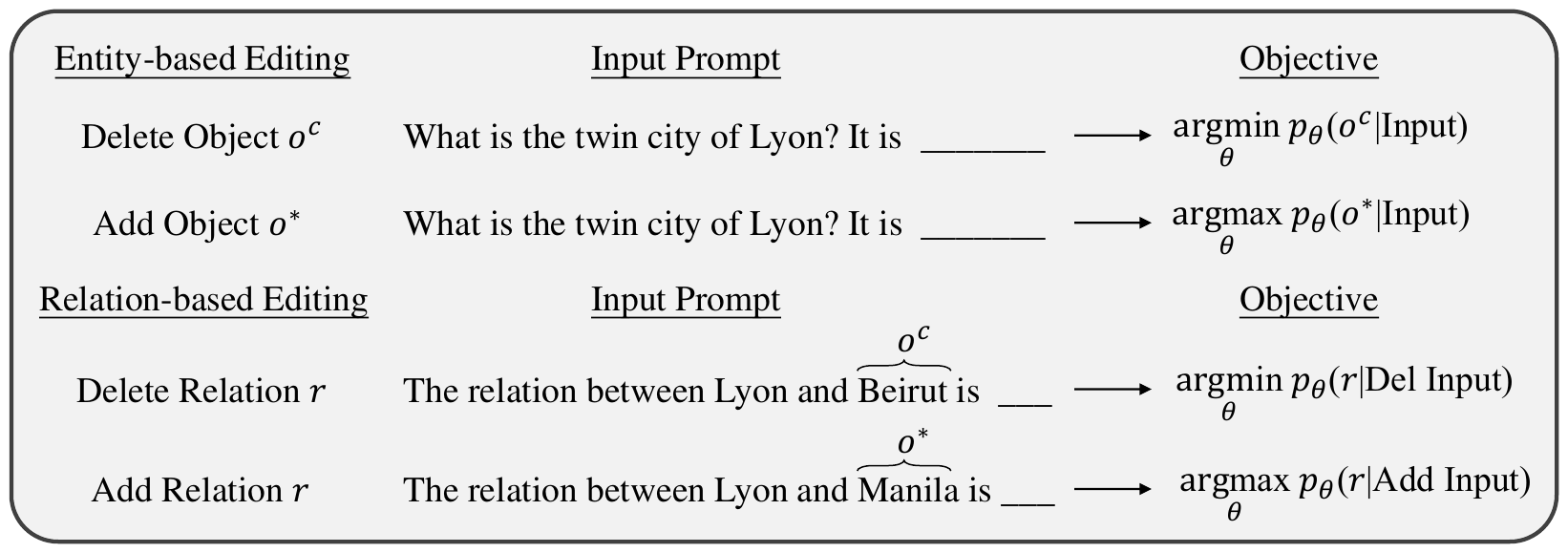}
        \end{center}
        \caption{\label{fig:Editing_problem_variants}
        Depiction of editing problem variants, where $r$ represents the relation P190 "twin city," $o^c$ and $o^*$ respectively represent the original object and the new object after editing.
        We can establish the logical equivalence of the editing results from both perspectives: instead of modifying a new object fact within the model (Entity-based Editing), we consider directly modifying the relation output (Relation-based Editing). 
       }
        \end{figure*}

    In this section, we will illustrate the proposed relation-based editing task in Figure \ref{fig:Editing_problem_variants}. We will discuss the task definition (§\ref{sec:task definition}), and explain the evaluation metrics (§\ref{sec:evaluation metrics}).

    \subsection{Task Definition}
    \label{sec:task definition}
    Following the work of \citep{petroni2019language}, we adopt the definition that a large language model possesses knowledge of a fact $P$ in the form of $(s,r,o)$. 
    In this context,  $s$ represents a subject entity (e.g., Lyon), $r$ represents a relation (e.g., twin city), and $o$ represents an object (e.g., Beirut). 
    We also use a few variations of the data for the fact $(s, r, o)$. The additional variables include:
    \begin{enumerate}
      \item $s^*$ represents a neighboring entity to the subject $s$ 
      (e.g. \textit{“Cairo”} is a neighboring entity to \textit{“Lyon”}), for which $(s^*, r, o)$ is a true fact like $(s, r, o)$. 
      \item $r^*$ is a paraphrase of the relation $r$ between the subject $s$ and object $o$, such as \textit{“[s] works in the field of [o]”} for \textit{“[s] works in the area of [o].”}
      \item $o^c$ is the original object that correctly completes the fact 
      $(s, r, \cdot)$, and $o^*$ is a new object after editing updates.
    \end{enumerate}
    As show in Figure \ref{fig:Editing_problem_variants}, we can establish the logical equivalence of the factual knowledge $P$ between entity perspective and relation perspective.
    In this paper, we propose  that the fact $P$ signifies the natural language prompt “The relation between Lyon and Beirut is \_\_\_” where the relation $r$ needs to be completed. 
    The main objective of the model editing task is to modify a base model $f_{\theta}$, parameterized by $\theta$, to gain control over the model's prediction outputs.
    Specifically, the base model $f_{\theta}$ is represented by a function $f: \mathbb{X} \mapsto \mathbb{Y}$
    that associates an input $P$ with its corresponding prediction $r$, as show in Equation \ref{equation1}.

    \begin{equation}
    \label{equation1}
    f_{\theta}(P)=
    \begin{cases}
    \underset{\theta}{\operatorname{argmax}} p_\theta(r\mid s,o) & \text{if } o \in o^* \\
    \underset{\theta}{\operatorname{argmin}} p_\theta(r\mid s,o) & \text{if } o \in o^c 
    \end{cases}
    \end{equation}

    To achieve control over the model's output, we aim for the model's conditional probability $p_{\theta}(r|s,o^{*})$ to be maximized and $p_{\theta}(r|s,o^{c})$ to be minimized. Here, $o^{c}$ represents the original tail entity, and $o^{*}$ represents the modified tail entity.

    \subsection{Evaluation Metrics}
    \label{sec:evaluation metrics}
    Model editing methods are commonly evaluated according to three aspects: 
    Efficacy: their effectiveness in altering the model prediction for the input prompt $P$. Generalization: generalize to paraphrases of the prompt $P$.
    Specificity: avoid side effects on irrelevant fact knowledge.
    
    In particular, we gather a set of more difficult false facts $(s,r,o^*)$, 
    these counterfactuals start with low scores compared to the correct facts $(s,r,o^c)$.
    Our editing objective is to establish a relationship $r$ between $s$ and $o^*$ while severing the connection $r$ between $s$ and $o^c$.
    To assess the efficacy of changes about relation, we divide the evaluation metrics into two: Success and Magnitude.
    The Success is the proportion of cases for which we have $p(r^*)>p(r^c)$ (or $p(o^*)>p(o^c)$) post-edit,
    and Magnitude is the average difference $p(r^*)-p(r^c)$ 
    (or $p(o^*)-p(o^c)$).
    In details, we report Efficacy Success (ES) and Efficacy Magnitude (EM) to  assess the efficacy of changes about relation,  we collect a set of rephrased prompts equivalent to $P$ and report Paraphrase Scores (PS) and (PM), we collect a set of nearby subjects $s_n$ for which $(s_n,r,o^c)$ holds true to measure Neighborhood Score NS and NM , computed similarly to ES and EM. To test three metrics tradeoff, we report the harmonic mean of ES, PS, NS as Score (S).

\section{RaKE: Relation-based Knowledge Editing}

A factual knowledge can be represented by a triplet $(s, r, o)$. In the entity perspective, the current approach predicts the object based on the given prompt $(s, r)$. In the relation perspective, it is equivalent to completing the relationship between the subject and object given $(s, o)$.
For example, “What is the twin city of Lyon? It is \_\_\_”, for which the expected completion is o = “Beirut”.
Equivalent to: “The relation between Lyon and Beirut is \_\_\_”, for which the expected completion is r = “twin city”.
To evaluate the editing capability of the current editing method for relation knowledge, we follow the dataset COUNTERFACT 
\citep{meng2022locating} and construct an equivalent relation perspective dataset named RaKE. We first present the data construction process for the dataset. Then, we present the data statistics
and evaluation settings of the RaKE, followed by evaluation metrics in the end.

\begin{table*}[ht]
\resizebox{1.0\linewidth}{!}
{
\begin{tabular}{lccccccc}
\specialrule{0.07em}{0pt}{0pt}
\textbf{Criterion} & zsRE & PARAREL & COUNTERFACT & Calibration & MQuAKE & RIPPLEEDITS & RaKE \\
Entity Efficacy     & \cmark & \cmark & \cmark & \cmark & \cmark & \cmark & \cmark \\
Entity Paraphrase   & \xmark & \cmark & \cmark & \cmark & \cmark & \cmark & \cmark \\
Specificity         & \xmark & \xmark & \cmark & \xmark & \cmark & \cmark & \cmark \\
Multi-hop           & \xmark & \xmark & \xmark & \xmark & \cmark & \xmark & \xmark \\
Relation Efficacy   & \xmark & \xmark & \xmark & \xmark & \xmark & \xmark & \cmark \\
Relation Paraphrase & \xmark & \xmark & \xmark & \xmark & \xmark & \xmark & \cmark \\
\specialrule{0.07em}{0pt}{0pt}
\end{tabular}
}
\caption{\label{tab:compariosn}Comparison to Existing Benchmarks. While previous benchmarks have defined factual knowledge in the form of triples $(s, r, o)$, 
existing paradigms assess whether an "entity-based" edit $(s, r \to o^*)$ is successful, but lack evaluation for the equivalent knowledge $(s, o^* \to r)$.
}
\end{table*}

\subsection{Dataset Construction}
\noindent \textbf{Generalization Dataset Construction. }
To compare and assess semantic generalization of the language model in the relation perspective, we collect relations based on Wikidata \citep{vrandevcic2014wikidata}, a knowledge base consisting of fact triples associated with thousands of relations.
We first manually select 34 common relations from wikidata 
and then leverage the PARAREL dataset \citep{elazar2021measuring} to get paraphrase for relations.  
Finally, we construct relation paraphrase prompts using manually designed templates, such as:
“When it comes to {subject} and {object}, the relation can be defined as \_\_\_”.
We also adopt GPT3.5-turbo model to ensure that the sampled fact triples are coherent and lead to natural questions about relations, such as:
“What is the correlation between Danielle Darrieux and English?”.

\noindent \textbf{Efficay Dataset Construction. }
In this paper, we define the knowledge editing task from a relational perspective using two atomic operations.
1) Delete operation: Removing the relation $r$ between $s$ and $o$.
2) Add operation: Adding the relation $r$ between $s$ and $o^*$, as illustrated in Figure \ref{fig:Editing_problem_variants}. 
By utilizing these two atomic operations, we have achieved the logical equivalence to the entity-based editing method.
We manually designed templates for these two atomic operations and constructed efficacy prompts for all facts by filling the slots.

\subsection{Dataset Comparison}
Table \ref{tab:compariosn} shows a comprehensive comparison of related datasets. RaKE is the first dataset to study relation-based knowledge editing over language models.
Due to the fact that factual knowledge is composed of tuple $(s,r,o)$, any change in one of these components will result in a transformation of the associated knowledge. Therefore, for the expression of the same factual knowledge, there are two perspectives: the relation perspective and the entity perspective. There exists a mutual dependence and feedback relationship between these two perspectives.
Compared with previous benchmarks, RaKE takes into account editing problem variants and incorporates evaluation prompts related to relation editing. It assesses the effectiveness of edits from a relational perspective, rather than solely measuring the accuracy of predicting the tail entity.

\subsection{Dataset Statistics}
The RaKE dataset consists of 21,919 editing samples, each of which can be categorized as either entity-based or relation-based. Each sample includes editing prompts for modifying knowledge, as well as Paraphrase Prompts and Neighborhood Prompts. Specifically, the entity-based category contains 21,919 Edit Prompts, 82,650 Neighborhood Prompts, and 43,838 Paraphrase Prompts. The relation-based category includes 43,838 Edit Prompts, 284,102 Paraphrase Prompts. Both categories share the entity-based Neighborhood Prompts to assess the impact on unrelated knowledge.
The dataset statitics are summarized in Table \ref{statistics}.

\begin{table}[ht]
\centering
\begin{tabular}{@{}lcc@{}}
\toprule
Type                 & $N_{Entity}$  & $N_{Relation}$   \\ \midrule
Edit Prompts         & 21919 & 43838  \\
Neighborhood Prompts & 82650 & -     \\
Paraphrase Prompts   & 43838 & 284102 \\ \bottomrule
\end{tabular}
\caption{\label{statistics} Statistics of RaKE. $N_{Entity}$ and $N_{Relation}$ represent the number of samples in the entity perspective and relation perspective, respectively.
}
\end{table}


\section{Experiments}
\label{sec:experiments}

In this section, we compare the performance differences between entity-based editing and relation-based editing and identify weaknesses in LLMs with respect to editing relations.
The results of these comparisons are displayed in Table \ref{tab:results}.
Furthermore, we analyze the storage and recall of relation memory in LLMs through Casual Tracing, as show in Figure \ref{fig:AIE}.

\begin{table*}[]
\resizebox{1.0\linewidth}{!}{
\begin{tabular}{ccccccccccccc}
\specialrule{0.1em}{0pt}{0pt}
\multicolumn{2}{c|}{\multirow{2}{*}{Editor}}                         & \multicolumn{1}{c|}{Score} & \multicolumn{2}{c|}{E-Efficacy}        & \multicolumn{2}{c|}{R-Efficacy}  & \multicolumn{2}{c|}{Specificity} & \multicolumn{2}{c|}{E-Generalization} 
& \multicolumn{2}{c}{R-Generalization} 
\\ \cline{3-13} 
\multicolumn{2}{c|}{} & \multicolumn{1}{c|}{S $\uparrow$} & \multicolumn{1}{c|}{ES $\uparrow$}    & \multicolumn{1}{c|}{EM $\uparrow$}    & \multicolumn{1}{c|}{ES $\uparrow$}    & \multicolumn{1}{c|}{EM $\uparrow$}    & \multicolumn{1}{c|}{NS $\uparrow$}    & \multicolumn{1}{c|}{NM $\uparrow$}    & \multicolumn{1}{c|}{PS $\uparrow$}    & \multicolumn{1}{c|}{PM $\uparrow$}    & \multicolumn{1}{c|}{PS $\uparrow$}      & PM $\uparrow$    
\\ \specialrule{0.06em}{0pt}{0pt}
\multicolumn{13}{c}{\coloritt Entity Perspective}   
\\ \specialrule{0.06em}{0pt}{0pt}
\multicolumn{1}{c|}{\multirow{5}{*}{\begin{tabular}[c]{@{}c@{}}GPT-2 \\ XL\end{tabular}}} 
& \multicolumn{1}{c|}{FT}    & \multicolumn{1}{c|}{72.98} & \multicolumn{1}{c|}{99.28} & \multicolumn{1}{c|}{92.1}  & \multicolumn{1}{c|}{97.19} & \multicolumn{1}{c|}{0.12}  & \multicolumn{1}{c|}{70.06} & \multicolumn{1}{c|}{3.6}   & \multicolumn{1}{c|}{48.21} & \multicolumn{1}{c|}{0.38}  & \multicolumn{1}{c|}{76.14}   & 0.09   \\ \cline{2-13} 
\multicolumn{1}{c|}{}      & \multicolumn{1}{c|}{KN}    & \multicolumn{1}{c|}{46.42} & \multicolumn{1}{c|}{30.45} & \multicolumn{1}{c|}{-2.08} & \multicolumn{1}{c|}{83.42} & \multicolumn{1}{c|}{0.08}  & \multicolumn{1}{c|}{69.19} & \multicolumn{1}{c|}{1.98}  & \multicolumn{1}{c|}{28.8}  & \multicolumn{1}{c|}{-1.92} & \multicolumn{1}{c|}{72.93}   & 0.05   \\ \cline{2-13} 

\multicolumn{1}{c|}{} & \multicolumn{1}{c|}{MEND}  
& \multicolumn{1}{c|}{67.81} & \multicolumn{1}{c|}{93.8}  & \multicolumn{1}{c|}{45.27} & \multicolumn{1}{c|}{97.91} & \multicolumn{1}{c|}{0.12}  & \multicolumn{1}{c|}{44.44} & \multicolumn{1}{c|}{-6.61} & \multicolumn{1}{c|}{58.0}  & \multicolumn{1}{c|}{7.88}  & \multicolumn{1}{c|}{76.22}   & 0.08   \\ \cline{2-13} 
\multicolumn{1}{c|}{}  & \multicolumn{1}{c|}{ROME}  & \multicolumn{1}{c|}{87.01} & \multicolumn{1}{c|}{99.93} & \multicolumn{1}{c|}{97.94} & \multicolumn{1}{c|}{96.12} & \multicolumn{1}{c|}{0.17}  & \multicolumn{1}{c|}{75.36} & \multicolumn{1}{c|}{4.4}   & \multicolumn{1}{c|}{96.6}  & \multicolumn{1}{c|}{62.91} & \multicolumn{1}{c|}{74.46}   & 0.09   \\ \cline{2-13} 
\multicolumn{1}{c|}{}    & \multicolumn{1}{c|}{MEMIT} & \multicolumn{1}{c|}{83.78} & \multicolumn{1}{c|}{93.88} & \multicolumn{1}{c|}{64.06} & \multicolumn{1}{c|}{97.28} & \multicolumn{1}{c|}{0.13}  & \multicolumn{1}{c|}{76.75} & \multicolumn{1}{c|}{4.97}  & \multicolumn{1}{c|}{79.6}  & \multicolumn{1}{c|}{26.24} & \multicolumn{1}{c|}{76.0}    & 0.09   \\ \hline \hline
\multicolumn{1}{c|}{\multirow{3}{*}{GPT-J}}    
& \multicolumn{1}{c|}{MEND}  & \multicolumn{1}{c|}{69.0}  & \multicolumn{1}{c|}{97.43} & \multicolumn{1}{c|}{72.12} & \multicolumn{1}{c|}{91.91} & \multicolumn{1}{c|}{0.11}  & \multicolumn{1}{c|}{53.15} & \multicolumn{1}{c|}{-5.44} & \multicolumn{1}{c|}{53.53} & \multicolumn{1}{c|}{11.12} & \multicolumn{1}{c|}{72.34}   & 0.08   \\ \cline{2-13} 
\multicolumn{1}{c|}{}     & \multicolumn{1}{c|}{ROME}  & \multicolumn{1}{c|}{87.51} & \multicolumn{1}{c|}{99.99} & \multicolumn{1}{c|}{99.49} & \multicolumn{1}{c|}{91.37} & \multicolumn{1}{c|}{0.13}  & \multicolumn{1}{c|}{78.61} & \multicolumn{1}{c|}{5.3}   & \multicolumn{1}{c|}{99.49} & \multicolumn{1}{c|}{77.21} & \multicolumn{1}{c|}{74.52}   & 0.09   \\ \cline{2-13} 
\multicolumn{1}{c|}{}  & \multicolumn{1}{c|}{MEMIT} 
& \multicolumn{1}{c|}{}      & \multicolumn{1}{c|}{}      & \multicolumn{1}{c|}{}      & \multicolumn{1}{c|}{}      & \multicolumn{1}{c|}{}      & \multicolumn{1}{c|}{}      & \multicolumn{1}{c|}{}      & \multicolumn{1}{c|}{}      & \multicolumn{1}{c|}{}      & \multicolumn{1}{c|}{}   &    
\\ \specialrule{0.06em}{0pt}{0pt}
\multicolumn{13}{c}{\coloritt Relation Perspective}                  \\ \specialrule{0.06em}{0pt}{0pt}
\multicolumn{1}{c|}{\multirow{5}{*}{\begin{tabular}[c]{@{}c@{}}GPT-2 \\ XL\end{tabular}}} 
& \multicolumn{1}{c|}{FT}    & \multicolumn{1}{c|}{42.76} & \multicolumn{1}{c|}{23.92} & \multicolumn{1}{c|}{-4.76} & \multicolumn{1}{c|}{98.79} & \multicolumn{1}{c|}{29.19} & \multicolumn{1}{c|}{76.69} & \multicolumn{1}{c|}{5.05}  & \multicolumn{1}{c|}{25.44} & \multicolumn{1}{c|}{-4.13} & \multicolumn{1}{c|}{79.03}   & 2.19   \\ \cline{2-13} 
\multicolumn{1}{c|}{} & \multicolumn{1}{c|}{KN}    
& \multicolumn{1}{c|}{41.23} & \multicolumn{1}{c|}{22.53} & \multicolumn{1}{c|}{-4.92} & \multicolumn{1}{c|}{97.52} & \multicolumn{1}{c|}{0.12}  & \multicolumn{1}{c|}{77.72} & \multicolumn{1}{c|}{5.17}  & \multicolumn{1}{c|}{24.61} & \multicolumn{1}{c|}{-4.09} & \multicolumn{1}{c|}{76.16}   & 0.08   \\ \cline{2-13} 
\multicolumn{1}{c|}{} & \multicolumn{1}{c|}{MEND}  
& \multicolumn{1}{c|}{41.57} & \multicolumn{1}{c|}{22.33} & \multicolumn{1}{c|}{-4.94} & \multicolumn{1}{c|}{100.0} & \multicolumn{1}{c|}{14.78} & \multicolumn{1}{c|}{77.63} & \multicolumn{1}{c|}{5.2}   & \multicolumn{1}{c|}{24.63} & \multicolumn{1}{c|}{-4.13} & \multicolumn{1}{c|}{83.24}   & 1.7    \\ \cline{2-13} 
\multicolumn{1}{c|}{} & \multicolumn{1}{c|}{ROME}  
& \multicolumn{1}{c|}{47.27} & \multicolumn{1}{c|}{27.92} & \multicolumn{1}{c|}{-3.7}  & \multicolumn{1}{c|}{99.99} & \multicolumn{1}{c|}{86.7}  & \multicolumn{1}{c|}{77.88} & \multicolumn{1}{c|}{5.09}  & \multicolumn{1}{c|}{28.12} & \multicolumn{1}{c|}{-3.76} & \multicolumn{1}{c|}{84.47}   & 15.16  \\ \cline{2-13} 
\multicolumn{1}{c|}{}     & \multicolumn{1}{c|}{MEMIT} & \multicolumn{1}{c|}{42.03} & \multicolumn{1}{c|}{24.15} & \multicolumn{1}{c|}{-4.11} & \multicolumn{1}{c|}{91.36} & \multicolumn{1}{c|}{3.84}  & \multicolumn{1}{c|}{77.66} & \multicolumn{1}{c|}{5.13}  & \multicolumn{1}{c|}{24.63} & \multicolumn{1}{c|}{-4.04} & \multicolumn{1}{c|}{76.24}   & 0.73   \\ \hline \hline
\multicolumn{1}{c|}{\multirow{3}{*}{GPT-J}}    
& \multicolumn{1}{c|}{MEND}  & \multicolumn{1}{c|}{32.38} & \multicolumn{1}{c|}{15.51} & \multicolumn{1}{c|}{-7.26} & \multicolumn{1}{c|}{100.0} & \multicolumn{1}{c|}{45.96} & \multicolumn{1}{c|}{82.77} & \multicolumn{1}{c|}{7.58}  & \multicolumn{1}{c|}{17.99} & \multicolumn{1}{c|}{-6.65} & \multicolumn{1}{c|}{81.52}   & 5.11   \\ \cline{2-13} 
\multicolumn{1}{c|}{}   & \multicolumn{1}{c|}{ROME}  
& \multicolumn{1}{c|}{51.98} & \multicolumn{1}{c|}{30.95} & \multicolumn{1}{c|}{-3.83} & \multicolumn{1}{c|}{100.0} & \multicolumn{1}{c|}{98.51} & \multicolumn{1}{c|}{82.75} & \multicolumn{1}{c|}{7.54}  & \multicolumn{1}{c|}{31.87} & \multicolumn{1}{c|}{-3.76} & \multicolumn{1}{c|}{95.97}   & 28.18  \\ \cline{2-13} 
\multicolumn{1}{c|}{}  & \multicolumn{1}{c|}{MEMIT} 
& \multicolumn{1}{c|}{}      & \multicolumn{1}{c|}{}      & \multicolumn{1}{c|}{}      & \multicolumn{1}{c|}{}      & \multicolumn{1}{c|}{}      & \multicolumn{1}{c|}{}      & \multicolumn{1}{c|}{}      & \multicolumn{1}{c|}{}      & \multicolumn{1}{c|}{}      & \multicolumn{1}{c|}{}        &        \\ \specialrule{0.1em}{0pt}{0pt}
\end{tabular}
}
\caption{\label{tab:results} 
The performance of knowledge editing approaches. In the table, R represents relation, and E represents Entity.
}
\end{table*}

\subsection{Experimental Setup}
\noindent \textbf{Language models. }
We use GPT-2 XL (1.5B) and GPT-J (6B) as the baseline models to assess model editing methods. In our experiment, we utilize four NVIDIA RTX A6000 GPUs and ten NVIDIA GeForce RTX 3090 GPUs to run model editing approaches.


\noindent \textbf{Model editing methods.}
In this paper, our focus is on transformer-based language models, specifically exploring the connection between model parameters and memory. Therefore, we employ memory-based and Locate-Then-Edit paradigms as our model editing methods.

\begin{itemize}
    \item \noindent 
    \textbf{Finetune. } Fine-tuning is a commonly used approach for adapting pre-trained language models to specific tasks or domains.In this paper, we compare with a naive fine-tuning approach that uses weight decay to prevent forgetfulness (FT).
    \item \noindent \textbf{KN. }  The Knowledge Neuron (KN) method \citep{dai2021knowledge}  introduces a knowledge attribution technique to identify the “knowledge neuron” (a key-value pair in the Feed-Forward Network matrix) encapsulate important memory. These neurons are then updated to incorporate relevant knowledge.
    \item \noindent \textbf{MEND. } Model Editor Networks with Gradient Decomposition \citep{mitchell2021fast} enables efficient local edits to language models by transforming the gradients of fine-tuned models. It achieves this by utilizing a low-rank decomposition of the gradients.
    \item \noindent \textbf{ROME. }  \citet{meng2022locating} applies causal mediation analysis to locate the specific areas requiring modifications. Instead of modifying individual knowledge neurons in the FFN, ROME iteratively updates one fact at a time by altering the entire matrix.
    \item \noindent \textbf{MEMIT. }  \citet{meng2022mass} is a method that allows for simultaneous modification of a sequence of layers in a language model. It facilitates thousands of alterations to be executed efficiently.
\end{itemize}

\subsection{Results and Analysis}
\noindent \textbf{Efficacy. } 
Entity-based editing focuses on completing the object given a prompt consisting of subject and relation, while relation-based editing involves completing the relationship between a subject and object given a prompt consisting of subject and object.
\begin{figure}[h]
    \centering
    \includegraphics[width=1\linewidth]{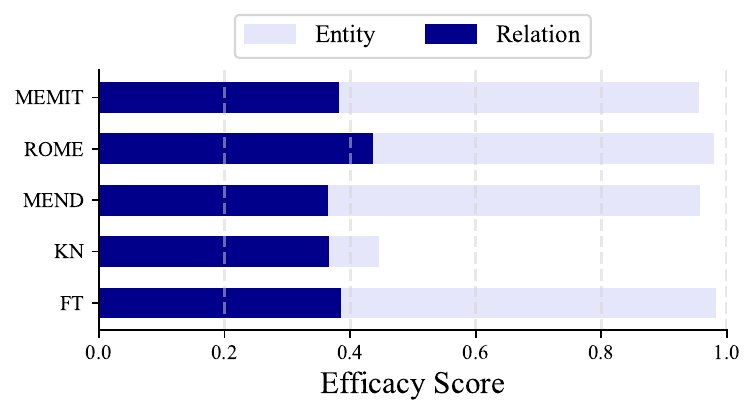}
    \caption{Efficacy Score}
    \label{fig:Efficacy_Score}
\end{figure}
We believe there is an equivalence relationship between the two. That is, modifying the tail entity is equivalent to adding a relation between the head entity and the tail entity.
However, according to Table \ref{tab:results}, we observe that current model editing methods are not applicable to relation perspective. Specifically, R-Efficacy shows a significant decrease in performance compared to E-Efficacy in terms of the EM metric.
This suggests that the existing editing methods, which work well for entity perspective tasks, do not effectively handle relation perspective tasks. There is a clear performance gap when it comes to editing relations, indicating the limitations of LLMs in accurately capturing and generating complex relationships between entities.
This finding highlights the need for further research and development of editing methods specifically tailored for relation perspective tasks, aiming to improve the performance and efficacy of LLMs in relation completion and understanding.

\noindent \textbf{Geralization. } 
We evaluate all methods on GPT-2 XL with knowledge edit in RaKE. The evaluation results are shown in Table \ref{tab:results}.
\begin{figure}[h]
    \centering
    \includegraphics[width=1\linewidth]{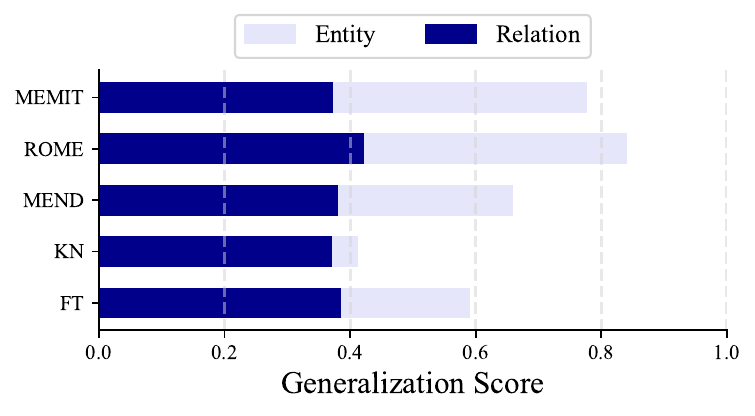}
    \caption{Generalization Score}
    \label{fig:Generalization_Score}
\end{figure}
From the results of the Entity based Generalization and Relation based Generalization metrics, we can conclude that both entity-based and relation-based methods improve the generalization within their respective perspectives. However, their impact on generalization from the other perspective is relatively limited. Despite the logical equivalence of the knowledge edited from the entity and relation perspectives in terms of triple representation, they exhibit surprising differences in effectiveness. This leads us to speculate that entity-based knowledge and relation-based knowledge are not equivalent in language models. Specifically, entity knowledge and relation knowledge demonstrate a certain level of independence and are stored in different parts of the model.

\subsection{Casual Tracing}
To explore the role of relations in factual triplets $(s,r,o)$ within model parameters, we need to analyze and identify the knowledge neurons that have the strongest causal effect on relations.
We utilized causal tracing for this purpose, involving three steps as follow:
\begin{itemize}
  \item \textbf{Clean run:} we pass a factual prompt $x$ into a model $f_{\theta}$ and collect all hidden activations 
  $\{h_i^{(l)}\ |\ i \in [1,T], l \in [1,L]\}$, where $T$ is number of input tokens and $L$ is number of layers within model $f_{\theta}$.
  \item \textbf{Corrupted run:} The relation embeddings are obfuscated from $f_{\theta}$ before the network runs,  after x is embedded as $[h_1^{(0)}, h_2^{(0)}, \ldots ,h_T^{(0)}]$, we set $h_i^{(0)} := h_i^{(0)} + \epsilon $ for all indices $i$ that correspond to the relation, where $ \epsilon \sim \mathcal{N}(0, \nu)$\footnote{We select $\nu$  to be 3 times larger than the empirical standard deviation of embeddings.}, and then we get a set of corrupted activations $\{h_{i \ast}^{(l)}\ | \ i \in [1,T], l\in[1,L]\}$.

  \item \textbf{Corrupted-with-restoration run:} Let $f_{\theta}$ runs computations on the noisy embeddings as in the corrupted baseline, except at some token $\hat{i}$ and layer $\hat{l}$.
  There, we hook $f_{\theta}$ so that it is forced to output the clean state $h^{(l)}_{\hat{i}}$, and future computations execute without further intervention.
\end{itemize}

\begin{figure*}
    \centering
    \includegraphics[width=1\linewidth]{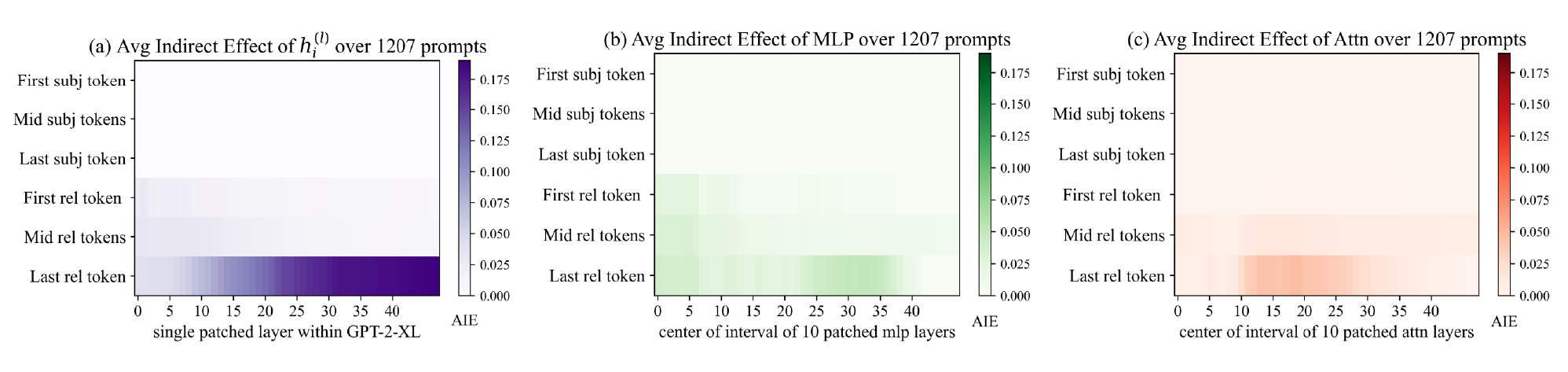}
    \caption{
    Causal tracing results of individual model components. In this paper, we use a sample of 1207 factual statements from \citep{meng2022locating} as knowledge queries to explore the knowledge contained within GPT-2 XL (1.5B).
    }
    \label{fig:AIE}
\end{figure*}

In our settings, $\mathbb{P}[r]$,  $\mathbb{P_*}[r]$, and  $\mathbb{P}_{*, \operatorname{clean} h_i^{(l)}}[r]$ is defined as the  probability of final prediction $r$ under the clean, corrupted, and corrupted-with-restoration runs, respectively. The indirect effect (IE) of a particular hidden state $h_i^l$ is calculated as:

\begin{equation}
\mathrm{IE}=\mathbb{P}_{*, \operatorname{clean} h_i^{(l)}}[r]-\mathbb{P}_*[r]
\end{equation}
IE is defined as the difference between the probability of $r$ under the corrupted version and the probability when that state is set to its clean version, while the relation remains corrupted. After averaging over all the prompts, we get the average average indirect effect (AIE) for each hidden state.

The result of the causal tracing analysis is depicted in Figure \ref{fig:AIE}. Consistently with previous findings, we observed a high AIE in the later layers of the final token. This implies that restoring the hidden states of the MLPs in those layers recovers most of the necessary information. Additionally, we noted a significant AIE in the earlier layers for the relation tokens that we intentionally corrupted. This discovery is non-trivial and underscores the significance of the earlier layers in predicting plausibility.
Furthermore, we observed a pronounced AIE in the middle attention layers of the last corrupted token. This surprising new finding suggests that memory related to relations is not only stored in the MLPs but also in the attention layers. 
This extends the previous finding that emphasized the significance of the attention module specifically at late site.

\subsection{Severed Causal Analysis}
To obtain a clearer understanding of the impact of MLP and Attn layers, we perform Severed causal tracing analysis with a modified causal graph, again following the footsteps of ROME.

\begin{figure}
    \centering
    \includegraphics[width=1\linewidth]{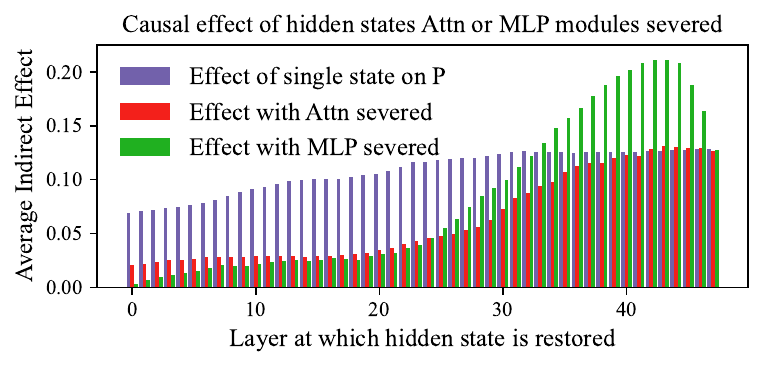}
    \caption{
    Causal effects with a modified GPT-2 XL model. To isolate the effects of individual model components, GPT-2 XL is modified by severing MLP and Attention modules. }
    \label{fig:AIE_for_each_module}
\end{figure}

Figure \ref{fig:AIE_for_each_module} presents a comparison of the average Average Individual Effect (AIE) at the last corrupted token for unmodified, severed MLP, and severed Attention causal graphs. Notably, we observe a distinct difference in AIE between the unmodified and severed MLP graphs, particularly in the earlier and middle layers. This finding is consistent with previous research and reinforces the critical role of MLP layers in plausibility prediction. Interestingly, the restoration effect appears to be independent of MLP activity in the higher layers, suggesting that the higher MLP layers may potentially generate unintended side effects.
In addition, we observe that the presence or absence of interruptions between attention modules in the range of 10 to 20 layers leads to significant differences in Attention-based Information Extraction (AIE). This finding suggests a strong correlation between the attention modules at layers 10 to 20 and relations. Consequently, we conclude that these parameters play a role in storing memory related to relations.

\subsection{Discussion}
Since a factual knowledge is composed of a subject, a relation, and an object, any change in one component will lead to a corresponding change in the related knowledge. Therefore, methods for knowledge editing should strive to achieve high success rates from both the relation and entity perspectives for the same factual knowledge. However, our experiments have shown that current knowledge editing methods face significant challenges and cannot simultaneously address both perspectives of a factual knowledge. Hence, future researchers in knowledge editing should explore the correlation and differences between entity and relation knowledge within language models, and fill the gaps in relation-based editing methods. Additionally, our tool has provided insights for future researchers, emphasizing the importance of not only focusing on the feed-forward networks (FFN) but also investigating the attention layer in editing tasks.
\section{Related Work}
\label{sec:bibtex}

\subsection{Memory in LLMs}
LLMs trained on extensive corpora such as Wikipedia, are widely believed to encapsulate vast amounts of factual knowledge \citep{petroni2019language, jiang2020can}.

\noindent\textbf{{FFN Memory. }}
\citet{geva2020transformer, geva2022transformer} show that feed-forward layers in transformer-based language models operate as key-value memories, where each key correlates with textual patterns in the training examples, and each value induces a distribution over the output vocabulary. The values complement the keys’ input patterns by inducing output distributions that concentrate probability mass on tokens likely to appear immediately after each pattern.
\citet{dai2021knowledge}  
proposes an attribution method to identify knowledge neurons that express factual knowledge in the FFN of pre-trained Transformers. They find that suppressing or amplifying the activation of knowledge neurons can accordingly affect the strength of knowledge expression.
\citet{meng2022locating, meng2022mass} develop a causal intervention for identifying neuron activations that are decisive in a model’s factual predictions. and then reveals an important role for mid-layer feed-forward modules that mediate factual predictions while processing subject tokens.

\noindent\textbf{{Attention Memory. }}
Sparse Distributed Memory (SDM) provides an efficient algorithm for storing and retrieving memories (patterns) from brain neurons. It effectively solves the "Best Match Problem" by quickly identifying the most suitable memory match for a given query.
Currently, \citet{bricken2021attention} has shown that the update rule of the Attention module in Transformer models closely approximates SDM.
Specifically, SDM consists of read and write operations. In the write operation, we can consider patterns being written and stored into nearby neurons based on the Hamming distance between patterns and neurons. In the read operation, the query is read from nearby neurons based on the circular region encompassed by the radius of the Hamming distance.
\citet{sakarvadia2023memory} establish an algorithm for injecting “memories” directly into the model’s hidden activations during inference. Through experimentation, they find that injecting relevant memories into the hidden activations of the attention heads during inference is an efficient way to boost model performance on multi-hop prompts.

\subsection{Memory Editing}

As factual information continues to evolve, the knowledge stored in LLMs can become outdated or incorrect. Hence, there is an urgent need to facilitate timely updates of inappropriate knowledge in LLMs while preserving other valuable knowledge. Recently, this issue has garnered significant attention from researchers.
Certainly, both parameter-efficient fine-tuning and incremental learning techniques provide avenues for modifying LLMs. However, it's essential to note that these approaches may be prone to overfitting and can incur substantial computational costs, especially when applied to large language models (LLMs) with an extremely large parameter scale.
To address these issues, \citet{sinitsin2020editable} proposes Model Editing, which aims to efficiently and accurately alter the factual knowledge stored within models.  Presently, there are three primary types of model editing approaches: 
1) Memory-based Method: These techniques utilize an additional trainable parameters to store memory or learn the required adjustments ($\Delta$) for knowledge updating in the LLMs \citep{de2021editing, mitchell2021fast, mitchell2022memory, dong2022calibrating, huang2023transformer}. 
2) Locate-Then-Edit Method: These approaches employ causal mediation analysis to locate knowledge neurons in LLMs and subsequently modify these recognized regions \citep{dai2021knowledge, meng2022locating, meng2022mass}.
3) In-Context Knowledge Editing Method: These methods are a training-free paradigm where knowledge editing is achieved directly by concatenating demonstrations within the input context \citep{zheng2023can, zhong2023mquake}.

\section{Conclusion}
In this paper, we introduce relation-based knowledge editing, with a new benchmark named \textbf{RaKE}. 
Empirically, we analyze the effectiveness of various model editing baselines and  notice that existing knowledge editing methods exhibit the potential difficulty in their ability to edit relations.
To investigate the fundamental reasons behind these results, we conducted causal analysis on the relationships within the triplets. We discovere that relational knowledge is not only stored in the FFN but also in the attention layer, which is a novel finding. From this, our experimental results indicate that the current editing methods, which focus solely on editing the parameters of the FFN module, lack modifications to the attention module. This inadequacy leads to suboptimal results in relation-based editing.

\section*{Limitations}
The current version of the RaKE dataset lacks an assessment of relation specificity performance, which we plan to include in future versions for evaluation. Furthermore, this is the first paper on relation perspective knowledge editing, and we acknowledge the lack of specific methods for editing relations. Our research serves as a preliminary investigation, and we will gradually refine the editing methods targeting relations in our subsequent work.




\bibliography{anthology,custom}

\begin{thebibliography}{25}
\expandafter\ifx\csname natexlab\endcsname\relax\def\natexlab#1{#1}\fi

\bibitem[{Bricken and Pehlevan(2021)}]{bricken2021attention}
Trenton Bricken and Cengiz Pehlevan. 2021.
\newblock Attention approximates sparse distributed memory.
\newblock \emph{Advances in Neural Information Processing Systems}, 34:15301--15315.

\bibitem[{Dai et~al.(2021)Dai, Dong, Hao, Sui, Chang, and Wei}]{dai2021knowledge}
Damai Dai, Li~Dong, Yaru Hao, Zhifang Sui, Baobao Chang, and Furu Wei. 2021.
\newblock Knowledge neurons in pretrained transformers.
\newblock \emph{arXiv preprint arXiv:2104.08696}.

\bibitem[{De~Cao et~al.(2021)De~Cao, Aziz, and Titov}]{de2021editing}
Nicola De~Cao, Wilker Aziz, and Ivan Titov. 2021.
\newblock Editing factual knowledge in language models.
\newblock \emph{arXiv preprint arXiv:2104.08164}.

\bibitem[{Dong et~al.(2022)Dong, Dai, Song, Xu, Sui, and Li}]{dong2022calibrating}
Qingxiu Dong, Damai Dai, Yifan Song, Jingjing Xu, Zhifang Sui, and Lei Li. 2022.
\newblock Calibrating factual knowledge in pretrained language models.
\newblock \emph{arXiv preprint arXiv:2210.03329}.

\bibitem[{Elazar et~al.(2021)Elazar, Kassner, Ravfogel, Ravichander, Hovy, Sch{\"u}tze, and Goldberg}]{elazar2021measuring}
Yanai Elazar, Nora Kassner, Shauli Ravfogel, Abhilasha Ravichander, Eduard Hovy, Hinrich Sch{\"u}tze, and Yoav Goldberg. 2021.
\newblock Measuring and improving consistency in pretrained language models.
\newblock \emph{Transactions of the Association for Computational Linguistics}, 9:1012--1031.

\bibitem[{Geva et~al.(2022)Geva, Caciularu, Wang, and Goldberg}]{geva2022transformer}
Mor Geva, Avi Caciularu, Kevin~Ro Wang, and Yoav Goldberg. 2022.
\newblock Transformer feed-forward layers build predictions by promoting concepts in the vocabulary space.
\newblock \emph{arXiv preprint arXiv:2203.14680}.

\bibitem[{Geva et~al.(2020)Geva, Schuster, Berant, and Levy}]{geva2020transformer}
Mor Geva, Roei Schuster, Jonathan Berant, and Omer Levy. 2020.
\newblock Transformer feed-forward layers are key-value memories.
\newblock \emph{arXiv preprint arXiv:2012.14913}.

\bibitem[{Guo et~al.(2022)Guo, Yang, and Abbasi}]{guo2022auto}
Yue Guo, Yi~Yang, and Ahmed Abbasi. 2022.
\newblock Auto-debias: Debiasing masked language models with automated biased prompts.
\newblock In \emph{Proceedings of the 60th Annual Meeting of the Association for Computational Linguistics (Volume 1: Long Papers)}, pages 1012--1023.

\bibitem[{Huang et~al.(2023)Huang, Shen, Zhang, Zhou, Rong, and Xiong}]{huang2023transformer}
Zeyu Huang, Yikang Shen, Xiaofeng Zhang, Jie Zhou, Wenge Rong, and Zhang Xiong. 2023.
\newblock Transformer-patcher: One mistake worth one neuron.
\newblock \emph{arXiv preprint arXiv:2301.09785}.

\bibitem[{Jiang et~al.(2020)Jiang, Xu, Araki, and Neubig}]{jiang2020can}
Zhengbao Jiang, Frank~F Xu, Jun Araki, and Graham Neubig. 2020.
\newblock How can we know what language models know?
\newblock \emph{Transactions of the Association for Computational Linguistics}, 8:423--438.

\bibitem[{Kasai et~al.(2022)Kasai, Sakaguchi, Takahashi, Bras, Asai, Yu, Radev, Smith, Choi, and Inui}]{kasai2022realtime}
Jungo Kasai, Keisuke Sakaguchi, Yoichi Takahashi, Ronan~Le Bras, Akari Asai, Xinyan Yu, Dragomir Radev, Noah~A Smith, Yejin Choi, and Kentaro Inui. 2022.
\newblock Realtime qa: What's the answer right now?
\newblock \emph{arXiv preprint arXiv:2207.13332}.

\bibitem[{Lei et~al.(2023)Lei, Luo, Yang, Liu, Liu, Lei, Huang, Wei, He, Zhao et~al.}]{lei2023tableqakit}
Fangyu Lei, Tongxu Luo, Pengqi Yang, Weihao Liu, Hanwen Liu, Jiahe Lei, Yiming Huang, Yifan Wei, Shizhu He, Jun Zhao, et~al. 2023.
\newblock Tableqakit: A comprehensive and practical toolkit for table-based question answering.
\newblock \emph{arXiv preprint arXiv:2310.15075}.

\bibitem[{Ma et~al.(2023)Ma, Xu, Wei, Chen, Wang, Liu, and Wu}]{ma2023ex}
Huanhuan Ma, Weizhi Xu, Yifan Wei, Liuji Chen, Liang Wang, Qiang Liu, and Shu Wu. 2023.
\newblock Ex-fever: A dataset for multi-hop explainable fact verification.
\newblock \emph{arXiv preprint arXiv:2310.09754}.

\bibitem[{Meng et~al.(2022{\natexlab{a}})Meng, Bau, Andonian, and Belinkov}]{meng2022locating}
Kevin Meng, David Bau, Alex Andonian, and Yonatan Belinkov. 2022{\natexlab{a}}.
\newblock Locating and editing factual associations in gpt.
\newblock \emph{Advances in Neural Information Processing Systems}, 35:17359--17372.

\bibitem[{Meng et~al.(2022{\natexlab{b}})Meng, Sharma, Andonian, Belinkov, and Bau}]{meng2022mass}
Kevin Meng, Arnab~Sen Sharma, Alex Andonian, Yonatan Belinkov, and David Bau. 2022{\natexlab{b}}.
\newblock Mass-editing memory in a transformer.
\newblock \emph{arXiv preprint arXiv:2210.07229}.

\bibitem[{Mitchell et~al.(2021)Mitchell, Lin, Bosselut, Finn, and Manning}]{mitchell2021fast}
Eric Mitchell, Charles Lin, Antoine Bosselut, Chelsea Finn, and Christopher~D Manning. 2021.
\newblock Fast model editing at scale.
\newblock \emph{arXiv preprint arXiv:2110.11309}.

\bibitem[{Mitchell et~al.(2022)Mitchell, Lin, Bosselut, Manning, and Finn}]{mitchell2022memory}
Eric Mitchell, Charles Lin, Antoine Bosselut, Christopher~D Manning, and Chelsea Finn. 2022.
\newblock Memory-based model editing at scale.
\newblock In \emph{International Conference on Machine Learning}, pages 15817--15831. PMLR.

\bibitem[{Neeman et~al.(2022)Neeman, Aharoni, Honovich, Choshen, Szpektor, and Abend}]{neeman2022disentqa}
Ella Neeman, Roee Aharoni, Or~Honovich, Leshem Choshen, Idan Szpektor, and Omri Abend. 2022.
\newblock Disentqa: Disentangling parametric and contextual knowledge with counterfactual question answering.
\newblock \emph{arXiv preprint arXiv:2211.05655}.

\bibitem[{Petroni et~al.(2019)Petroni, Rockt{\"a}schel, Lewis, Bakhtin, Wu, Miller, and Riedel}]{petroni2019language}
Fabio Petroni, Tim Rockt{\"a}schel, Patrick Lewis, Anton Bakhtin, Yuxiang Wu, Alexander~H Miller, and Sebastian Riedel. 2019.
\newblock Language models as knowledge bases?
\newblock \emph{arXiv preprint arXiv:1909.01066}.

\bibitem[{Sakarvadia et~al.(2023)Sakarvadia, Ajith, Khan, Grzenda, Hudson, Bauer, Chard, and Foster}]{sakarvadia2023memory}
Mansi Sakarvadia, Aswathy Ajith, Arham Khan, Daniel Grzenda, Nathaniel Hudson, Andr{\'e} Bauer, Kyle Chard, and Ian Foster. 2023.
\newblock Memory injections: Correcting multi-hop reasoning failures during inference in transformer-based language models.
\newblock \emph{arXiv preprint arXiv:2309.05605}.

\bibitem[{Sinitsin et~al.(2020)Sinitsin, Plokhotnyuk, Pyrkin, Popov, and Babenko}]{sinitsin2020editable}
Anton Sinitsin, Vsevolod Plokhotnyuk, Dmitriy Pyrkin, Sergei Popov, and Artem Babenko. 2020.
\newblock Editable neural networks.
\newblock \emph{arXiv preprint arXiv:2004.00345}.

\bibitem[{Vrande{\v{c}}i{\'c} and Kr{\"o}tzsch(2014)}]{vrandevcic2014wikidata}
Denny Vrande{\v{c}}i{\'c} and Markus Kr{\"o}tzsch. 2014.
\newblock Wikidata: a free collaborative knowledgebase.
\newblock \emph{Communications of the ACM}, 57(10):78--85.

\bibitem[{Wei et~al.(2023)Wei, Su, Ma, Yu, Lei, Zhang, Zhao, and Liu}]{wei2023menatqa}
Yifan Wei, Yisong Su, Huanhuan Ma, Xiaoyan Yu, Fangyu Lei, Yuanzhe Zhang, Jun Zhao, and Kang Liu. 2023.
\newblock Menatqa: A new dataset for testing the temporal comprehension and reasoning abilities of large language models.
\newblock \emph{arXiv preprint arXiv:2310.05157}.

\bibitem[{Zheng et~al.(2023)Zheng, Li, Dong, Fan, Wu, Xu, and Chang}]{zheng2023can}
Ce~Zheng, Lei Li, Qingxiu Dong, Yuxuan Fan, Zhiyong Wu, Jingjing Xu, and Baobao Chang. 2023.
\newblock Can we edit factual knowledge by in-context learning?
\newblock \emph{arXiv preprint arXiv:2305.12740}.

\bibitem[{Zhong et~al.(2023)Zhong, Wu, Manning, Potts, and Chen}]{zhong2023mquake}
Zexuan Zhong, Zhengxuan Wu, Christopher~D Manning, Christopher Potts, and Danqi Chen. 2023.
\newblock Mquake: Assessing knowledge editing in language models via multi-hop questions.
\newblock \emph{arXiv preprint arXiv:2305.14795}.

\end{thebibliography}
\bibliographystyle{acl_natbib}

\appendix



\end{document}